\DocumentMetadata{lang=eng-US,pdfversion=1.7}
\documentclass[colorlinks,upint,subscriptcorrection,varvw,hyphenate,balance]{asmeconf}

\hypersetup{%
  pdfauthor={Nehal Afifi and Sven Matthiesen},
  pdftitle={A Physics-Informed Learning Framework for Linking Microscopic Surface Degradation to Bolted Joint Functional Behavior},
  pdfkeywords={corrosion, bolted joints, physics-informed neural networks, partial differential equation discovery, computer vision, circular engineering},
  pdfsubject={ASME IMECE conference paper}
}

\usepackage{amsmath,bm}
\usepackage{graphicx}
\usepackage{enumitem}
\usepackage{siunitx}
\usepackage{svg}
\allowdisplaybreaks

\newcommand{\vect}[1]{\bm{#1}}
\newcommand{\loss}{\mathcal{L}}

\newcommand{\surf}{\Gamma_s}
\newcommand{\diff}{\mathrm{d}}
\newcommand{\R}{\mathbb{R}}

\begin{document}

\ConfName{Proceedings of the ASME 2026\linebreak International Mechanical Engineering Congress and Exposition}
\ConfAcronym{IMECE2026}
\ConfDate{November 8--12, 2026}
\ConfCity{Vancouver, Canada}
\PaperNo{IMECE2026-192044}

\title{A Physics-Informed Learning Framework for Linking Microscopic Surface Degradation to Bolted Joint Functional Behavior}

\SetAuthors{%
  Nehal Afifi\affil{1}\CorrespondingAuthor{nehal.afifi@kit.edu},
  Gerhard Kachel\affil{2},
  Sven Matthiesen\affil{1}
}
\SetAffiliation{1}{Karlsruhe Institute of Technology (KIT), IPEK -- Institute of Product Engineering, Karlsruhe, Germany}
\SetAffiliation{2}{Offenburg University of Applied Sciences, Department of Mechanical and Process Engineering, Offenburg, Germany}
\maketitle


\begin{abstract}
Returned products in circular engineering often have heterogeneous degradation
histories, making remaining function difficult to assess. For bolted joints,
corrosion alters surface morphology, contact, friction, preload retention, and
mechanical response. This paper presents a physics-informed multimodal
framework for linking microscopic and camera-based degradation observations to
functional indicators of corroded bolted joints. The workflow defines the
corrosion failure mode, generates synchronized inspection and
functional-response records, preprocesses image, roughness, exposure, and
joint descriptors, aligns microscopy and camera observations through similarity
learning, and constructs a calibrated corrosion-severity field. Partial
differential equation (PDE) discovery is applied to this field to identify
candidate effective mesoscale laws rather than complete electrochemical
mechanisms. The corrosion-evolution model is coupled to a separate functional
model, with friction coefficient as the primary target and preload loss
included when clamp-force data are available.
The contribution is a reproducible pathway for physically constrained
assessment of degraded bolted-joint behavior.
\end{abstract}
\begin{nomenclature}
\EntryHeading{Roman letters}
\entry{$A_r$}{rust-area fraction}
\entry{$c(\vect{s},t)$}{calibrated corrosion-severity field}
\entry{$F^c$}{clamp-force response during tightening}
\entry{$F_p$}{preload or clamp-force condition}
\entry{$h_s$}{surface-height or roughness-related degradation field}
\entry{$I^m, I^c$}{microscopic and camera images}
\entry{$\vect{q}, \vect{e}$}{joint and exposure descriptors}
\entry{$\vect{x}^{\mathrm{cond}}, \vect{y}^{\mathrm{func}}$}{condition-input and functional-target vectors}
\entry{$\vect{z}$}{latent condition representation}
\EntryHeading{Greek letters}
\entry{$\mu$}{friction coefficient}
\entry{$\Delta F_p$}{preload or clamp-force loss}
\entry{$\sigma$}{stress-, preload-, or contact-related descriptor}
\entry{$\Theta(c)$}{candidate library for PDE discovery}
\entry{$\Gamma_s$}{inspected surface domain}
\entry{$\mathcal{R}_{\theta}$}{physics-informed PDE residual}
\end{nomenclature}

\section{INTRODUCTION}
Circular factories aim to keep products and components in use for as long as possible. This requires decisions about whether returned products  should
be reused, refurbished, disassembled, or recycled
\cite{potting2017circular,reike2018rs}. For bolted joints, these decisions are
difficult because the remaining function depends not only on the visible
condition of the joint, but also on preload, friction, contact condition,
tightening history, joint configuration, and the conditions of the thread and
bearing surfaces \cite{bickford2007bolted,buda2003bolted,vdi2230part3}.
Corrosion changes these surface and contact conditions by altering oxide
morphology, roughness, material loss, and local contact pressure
\cite{king2023review,albert2024chemical,jiang2022preload,afifi2025probabilistic}. The practical challenge is therefore to infer remaining functional behavior from inspection evidence that is feasible to collect from degraded products.\cite{afifi2024data}

Feasible inspection data for degraded bolted joints may include camera images,
microscopic images, roughness measurements, exposure information, and selected
functional measurements such as torque-angle or torque/clamp-force data
\cite{iso16047}. These data are informative but heterogeneous. They do not
directly provide a corrosion-state field, an effective degradation relation, or
a direct prediction of joint function. Existing methods address parts of this
problem: mechanistic corrosion models describe degradation physics, vision
methods extract visual descriptors, and physics-informed learning can impose
differential-equation constraints
\cite{martinezpaneda2024review,wu2025fusion,raissi2019pinn,rudy2017pdefind}.
What remains missing is a structured path that connects inspection evidence to
degradation-state representation and then to selected joint's functional indicators.

This paper presents a physics-informed multimodal learning framework for
linking microscopic and camera-based surface degradation observations to
selected bolted-joint functional behavior. The framework converts image,
roughness, exposure, and joint-design information into calibrated
corrosion-state fields, applies PDE discovery to these fields as candidate
effective mesoscale corrosion laws, and couples corrosion evolution to
functional prediction. The paper makes four contributions. First, it
formulates an inspection-oriented corrosion-state inference problem for
degraded bolted joints. Second, it defines a latent-field interface that
converts multimodal imaging and roughness observations into calibrated
corrosion-state fields. Third, it formulates genetic-algorithm-based PDE
discovery for candidate effective corrosion laws over those fields. Fourth, it
couples corrosion-state evolution to prediction of selected functional
indicators, with friction coefficient as the first target and preload loss as
a secondary target when clamp-force data are available.
\section{STATE OF THE ART}
Bolted joints are widely used detachable joining elements in mechanical
products \cite{bickford2007bolted,buda2003bolted}. Their behavior is governed
by preload, friction, contact condition, tightening procedure, joint
configuration, and the conditions of the thread and bearing surfaces. VDI 2230
and ISO 16047 provide the basis for relating joint loads, assembly conditions,
torque/clamp-force behavior, friction coefficient, and preload-related
quantities \cite{vdi2230part1,vdi2230part2,vdi2230part3,iso16047}. Preload
retention is also a central concern under loading \cite{junker1969}.

Corrosion changes these mechanisms by altering oxide morphology, roughness,
material loss, local contact pressure, and interfacial contact
\cite{king2023review,albert2024chemical}. Studies of mechanically fastened and
high-strength bolted joints show that corrosion can alter frictional behavior
and contribute to preload or clamping-force loss
\cite{saai2023galvanic,jiang2022preload,li2021clamping,shah2025guided}.
These findings motivate the central problem addressed in this paper: visible
and microscopic corrosion observations must be transformed into physically
meaningful information about selected functional indicators of the joint.

Circular engineering strengthens this need because returned products often have
unknown use histories, exposure conditions, maintenance histories, and
degradation paths. R-strategy decisions therefore require evidence about
remaining function and recoverability, not only visual severity
\cite{potting2017circular,reike2018rs}. Practical inspection of bolted joints
can provide camera images, microscopic images, roughness measurements, exposure
descriptors, and selected functional tests. The difficulty is that these
measurements live at different scales and in different data forms. Without a
common representation, they cannot be directly used for field-based degradation
modeling or functional prediction.

Mechanistic corrosion models describe transport, electrochemical kinetics,
pitting, and coupled degradation processes
\cite{mai2016phasefield,ansari2018phasefield}. Phase-field models have been
extended to pH-dependent corrosion, electro-chemo-mechanical coupling, and
microstructure-sensitive degradation
\cite{tsuyuki2018ph,cui2023ecm,makuch2024micro}. However, these models
typically require state variables, parameters, and boundary conditions that are
not available in inspection of returned products
\cite{king2023review,martinezpaneda2024review}. Computer-vision methods can
extract information from images at scale, but many corrosion applications
remain focused on detection, segmentation, or severity rating
\cite{wu2025fusion,vangrunderbeek2025field}. These outputs are useful, but
they do not by themselves explain how degradation changes bolted-joint
function.

Physics-informed neural networks (PINNs) provide a mechanism for combining
data with differential-equation constraints \cite{raissi2019pinn}. Data-driven PDE
discovery can identify candidate differential equations from measured or
reconstructed fields when the dependent variable and time axis are well
defined \cite{raissi2017discovery,rudy2017pdefind}, and weak-form
identification improves robustness when derivative estimates are noisy
\cite{messenger2021wsindy,afifi2026gaussian}. Multimodal representation learning and
registration provide tools for relating microscopic and camera-scale
observations in a common surface frame
\cite{baltrusaitis2019multimodal,pielawski2020comir,sun2021loftr}. The
novelty of the present work is not the invention of each individual module:
U-Net, contrastive learning, PDE-FIND, genetic symbolic search, and PINNs are
established methods
\cite{ronneberger2015unet,chen2020simclr,rudy2017pdefind,chen2022sga,raissi2019pinn,afifi2025data}.
This motivates the system-level formulation developed in this paper. In this
formulation, image and roughness measurements are transformed into calibrated
surface fields, PDE discovery is applied to those fields rather than directly
to pixels, and the resulting candidate corrosion evolution is coupled to
functional indicators that can support inspection decisions.

\section{PHYSICS-INFORMED MULTIMODAL FRAMEWORK}
\begin{figure*}[t]
    \centering
    \includegraphics[width=1.0\textwidth, trim=1cm 20.6cm 1cm 1cm, clip, keepaspectratio]{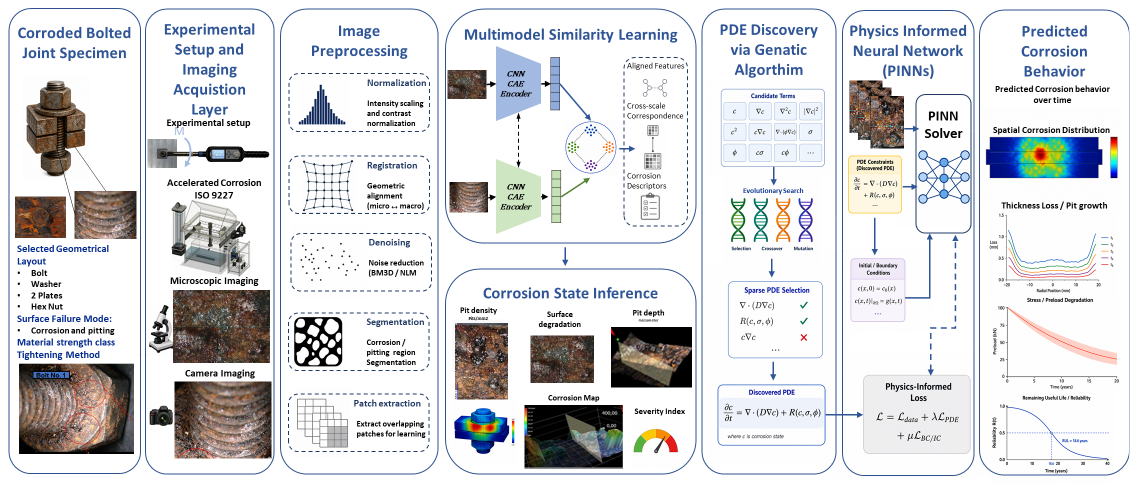}
    \caption{Overview of the proposed framework. The workflow connects problem definition, failure-mode selection, multimodal acquisition, preprocessing, similarity learning, corrosion-state inference, PDE discovery, physics-informed prediction, and circularity-oriented engineering interpretation.}
    \label{fig:overview}
\end{figure*}

Figure~\ref{fig:overview} shows the full framework. The figure is used as the organizing structure for the methodology. Each block produces an explicit output that becomes the input to the next block. This avoids treating the framework as a loose collection of modules.
The following subsections therefore move from system definition and data generation to representation learning, field construction, PDE discovery, prediction, and engineering interpretation.

\subsection{Problem Context and System of Interest}
The system of interest is a degraded bolted joint composed of a bolt, nut, washer, and clamped members. The inspected surface domain is denoted by $\surf\subset\R^2$, with surface coordinates $\vect{s}=(x,y)$. The joint is observed at exposure or inspection times $t_k$, where $k=0$ denotes the baseline condition and $k>0$ denotes post-exposure states. The modeling objective is to infer condition-related fields over $\surf$ and use them to predict selected functional indicators of the joint.

For specimen $i$ at time $t_k$, the synchronized multimodal record is defined as
\begin{equation}
\mathcal{O}_{i,k}=\left\{I^{m}_{i,k},I^{c}_{i,k},r_{i,k},\tau_{i,k}(\theta),F^{c}_{i,k}(\theta),F^{u}_{i,k}(u),\vect{q}_{i},\vect{e}_{i,k}\right\}
\label{eq:observations}
\end{equation}
In Eq.~\eqref{eq:observations}, $I^{m}_{i,k}$ is the microscopic image, $I^{c}_{i,k}$ is the camera or phone image, $r_{i,k}$ is a roughness or surface-profile measurement, $\tau_{i,k}(\theta)$ is the torque-angle response as a function of tightening angle $\theta$, $F^{c}_{i,k}(\theta)$ is the clamp-force response during tightening, $F^{u}_{i,k}(u)$ is an optional force-displacement response as a function of displacement $u$, $\vect{q}_{i}$ contains joint configuration variables such as bolt size and strength grade, and $\vect{e}_{i,k}$ contains exposure variables such as corrosion duration, medium, temperature, humidity, or chloride-related descriptors when measured. This record is then split into condition inputs and functional targets in the preprocessing step.

\subsection{Failure Mode Definition and Design of Experiments}
The target failure mode defines the data-generation space and the dependent degradation variable. Corrosion is selected as the governing degradation mechanism because it alters surface morphology, roughness, contact interface, friction, and preload retention. The design of experiments varies factors that influence both condition and response: corrosion exposure, preload level, joint configuration, bolt size, and strength grade. Statistical design-of-experiments principles are used to ensure that these factors span meaningful variability rather than a single narrow degradation path \cite{montgomery2017doe}.

The minimum reproducible data path requires repeated or re-localizable observations of the same physical regions. For each exposure/preload condition, multiple specimens are needed to separate specimen variability from exposure effects. Each specimen should include a baseline measurement and at least three post-exposure intervals if PDE discovery is intended. Each time point should include camera images, microscopy patches, roughness or profilometry data, and at least one functional measurement. The first functional target is the friction coefficient $\mu$ because it can be derived from torque/clamp-force testing and is directly linked to the mechanical behavior of bolted joints \cite{iso16047}.
Preload loss is introduced as a secondary target when clamp-force measurements are available.
These requirements define the experimental record that is described in the next steps.

\subsection{Multimodal Data Acquisition}
The acquisition step produces two groups of data. The first group contains condition-related observations: microscopic images, camera images, and roughness measurements. Microscopy captures local pits, oxide morphology, micro-scale roughness, and surface damage. Camera images capture larger-field visible corrosion, geometry, rust distribution, and spatial context. Roughness measurements provide scalar or profile-based descriptors such as $R_a$, $R_z$, and local height variation. The second group contains functional-response data, including torque-angle curves, torque/clamp-force relations, and optional force-displacement responses.
The acquisition step must preserve traceability. Each microscopic patch is linked to a specimen, time point, approximate camera-image region, roughness measurement location, and functional-response record. Without this traceability, image-derived fields cannot be related to functional behavior or used for PDE discovery. The output of this step is therefore not a set of independent images, but a synchronized multimodal record indexed by specimen, time, and surface location.

\subsection{Data Structuring and Preprocessing}
After acquisition, the synchronized record is converted into model-ready condition descriptors and functional targets. Camera images are corrected for illumination and scale, and the joint region of interest is segmented. Microscopic images are denoised, normalized, checked for focus quality, and divided into patches. Roughness profiles are converted into scalar and local descriptors. Functional curves are converted into interpretable behavior indicators.

The condition descriptor vector is defined as
\begin{equation}
\vect{x}^{\mathrm{cond}}_{i,k}=\left[\vect{d}^{\mathrm{img}}_{i,k},\vect{d}^{\mathrm{rough}}_{i,k},\vect{q}_{i},\vect{e}_{i,k}\right]
\label{eq:conditionvector}
\end{equation}
where $\vect{d}^{\mathrm{img}}_{i,k}$ contains image-derived descriptors such as rust coverage, pit-area fraction, pit-density estimates, and oxide texture features; $\vect{d}^{\mathrm{rough}}_{i,k}$ contains roughness descriptors such as $R_a$, $R_z$, local height variation, and texture directionality; $\vect{q}_{i}$ contains joint-design descriptors; and $\vect{e}_{i,k}$ contains exposure descriptors. The functional target vector is defined separately as
\begin{equation}
\vect{y}^{\mathrm{func}}_{i,k}=\left[\mu_{i,k},\Delta F_{p,i,k},\vect{a}^{\tau}_{i,k},\vect{a}^{F}_{i,k}\right]
\label{eq:functionvector}
\end{equation}
In Eq.~\eqref{eq:functionvector}, $\mu_{i,k}$ is the friction coefficient, $\Delta F_{p,i,k}$ is preload or clamp-force loss, $\vect{a}^{\tau}_{i,k}$ contains parameters extracted from the torque-angle curve, and $\vect{a}^{F}_{i,k}$ contains parameters extracted from the optional force-displacement curve. The clamp-force response supports the extraction of $\mu_{i,k}$ and $\Delta F_{p,i,k}$. The distinction between Eqs.~\eqref{eq:conditionvector} and \eqref{eq:functionvector} is essential: condition observations are model inputs, while functional variables are prediction targets.
This split provides the interface between inspection evidence and later functional prediction.

Segmentation and morphology extraction can be implemented with U-Net-like architectures or generalist microscopy segmentation tools when labeled data are available \cite{ronneberger2015unet,stringer2021cellpose}.
Image restoration may support low signal-to-noise ratio (SNR) microscopy, but restored images should not be treated as physical ground truth unless restoration artifacts are checked against raw observations \cite{weigert2018care,lehtinen2018noise2noise}.

\subsection{Multimodal Similarity Learning and Feature Fusion}
The next step addresses the scale mismatch between microscopic and camera-based observations. Microscopic and camera images represent the same surface degradation at different scales. The framework uses a two-branch representation model to align these observations in a shared latent space. Let $E_m$ be the microscopic encoder and $E_c$ be the camera-image encoder. Their patch embeddings are
\begin{equation}
\vect{h}^{m}_{i,k,j}=E_m(P^{m}_{i,k,j}),\qquad
\vect{h}^{c}_{i,k,j}=E_c(P^{c}_{i,k,j})
\label{eq:embeddings}
\end{equation}
where $P^{m}_{i,k,j}$ and $P^{c}_{i,k,j}$ are corresponding microscopic and camera-scale patches for specimen $i$, time $t_k$, and region $j$. The vectors $\vect{h}^{m}_{i,k,j}$ and $\vect{h}^{c}_{i,k,j}$ are the learned feature embeddings.
Cross-scale alignment is trained using a contrastive objective,
\begin{equation}
\loss_{\mathrm{sim}}=-\log\frac{\exp\left(\operatorname{sim}(\vect{h}^{m},\vect{h}^{c+})/T\right)}{\sum_{n}\exp\left(\operatorname{sim}(\vect{h}^{m},\vect{h}^{c}_{n})/T\right)}
\label{eq:simloss}
\end{equation}
where $\vect{h}^{c+}$ is the positive camera embedding paired with the microscopic embedding, $\vect{h}^{c}_{n}$ are negative or non-corresponding camera embeddings, $\operatorname{sim}(\cdot)$ is a cosine similarity function, and $T$ is the contrastive temperature. Siamese and contrastive learning provide the basis for this alignment \cite{bromley1993siamese,chen2020simclr}.
CoMIR-style shared representations and LoFTR-style local matching are suitable for dense cross-modal alignment after coarse pairing \cite{pielawski2020comir,sun2021loftr}.

\begin{figure*}[t]
    \centering
    \includegraphics[width=0.8\textwidth, trim=0.1cm 0cm 0cm 0cm, clip, keepaspectratio]{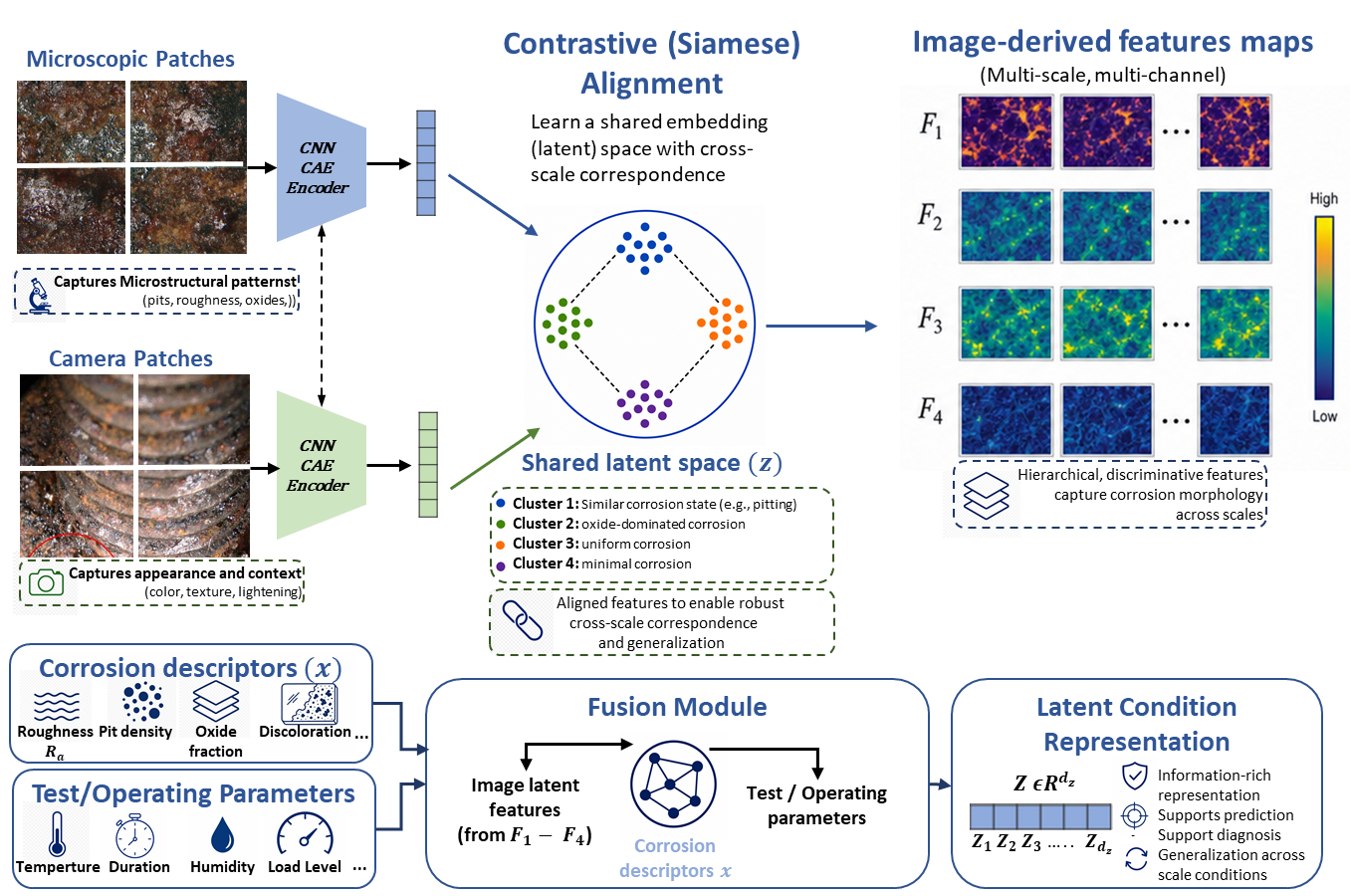}
    \caption{Cross-scale similarity learning and feature fusion. Microscopic and camera patches are embedded in a shared latent space and fused with roughness, exposure, and joint descriptors to produce a latent condition representation.}
    \label{fig:fusion}
\end{figure*}

The fused condition representation is written as
\begin{equation}
\vect{z}_{i,k,j}=F_{\omega}\left(\vect{h}^{m}_{i,k,j},\vect{h}^{c}_{i,k,j},\vect{d}^{\mathrm{rough}}_{i,k,j},\vect{q}_{i},\vect{e}_{i,k}\right)
\label{eq:fusion}
\end{equation}
where $F_{\omega}$ is the fusion model with parameters $\omega$, and $\vect{z}_{i,k,j}$ is the local latent condition representation. The local representations are registered to the inspected surface frame and aggregated or interpolated to obtain $\vect{z}_{i}(\vect{s},t_k)$ over $\surf$. Figure~\ref{fig:fusion} illustrates this similarity-learning and fusion step. Its output is not the final prediction; it is the structured representation used to reconstruct corrosion-state fields.

\subsection{Corrosion State Inference and Latent-Field Construction}
The fused representation is then converted into the field variable used by the physics-informed part of the framework. The primary degradation variable is defined as a calibrated corrosion-severity field,
\begin{equation}
    c_{i}(\vect{s},t_k)=C_{\eta}\left(m_{r},p,h_s,m_{\ell},\vect{z}_{i}(\vect{s},t_k);\vect{s},t_k\right),\quad c_i\in[0,1]
    \label{eq:cfield}
\end{equation}
In Eq.~\eqref{eq:cfield}, $c_i(\vect{s},t_k)$ is the corrosion-severity field for specimen $i$ at surface location $\vect{s}$ and time $t_k$; $C_{\eta}$ is the calibration model with parameters $\eta$; $m_r$ is a rust or oxide mask; $p$ is a pit-probability or pit-density field; $h_s$ is a surface-height or roughness-related degradation field; $m_{\ell}$ is a material-loss measurement when available; and $\vect{z}_{i}(\vect{s},t_k)$ is the fused latent condition representation in the surface frame. The field $c$ is therefore a calibrated surrogate. It is not a direct electrochemical state variable.

The calibration hierarchy is direct observables $\rightarrow$ reconstructed fields $\rightarrow$ fused surrogate $c$. Direct observables include images, roughness, microscopy, and optional material loss. Reconstructed fields include rust masks, pit maps, and roughness maps. The surrogate $c$ integrates these reconstructed fields and is then used for functional prediction. Functional targets such as $\mu$ or $\Delta F_p$ are kept separate so that the corrosion field does not simply encode the mechanical response.

The calibration loss is written as
\begin{equation}
\begin{aligned}
\loss_{c} ={}&
\lambda_r\|\hat{m}_{r}-m_{r}\|^2
+ \lambda_p\|\hat{p}-p\|^2 \\
&+ \lambda_h\|\hat{h}_{s}-h_{s}\|^2
+ \lambda_{\ell}\|\hat{m}_{\ell}-m_{\ell}\|^2
\end{aligned}
\label{eq:calibrationloss}
\end{equation}
where hatted quantities are auxiliary reconstructions produced during calibration, and $\lambda_r$, $\lambda_p$, $\lambda_h$, and $\lambda_\ell$ weight rust-mask, pit-field, surface-height, and material-loss calibration terms. If a measurement is not available, its term is omitted rather than imputed as ground truth. Equation~\eqref{eq:calibrationloss} ensures that $c$ remains connected to independent degradation evidence.

\subsection{PDE Discovery from Degradation Data}
Once calibrated corrosion fields are available at repeated inspection times, PDE discovery is applied to $c(\vect{s},t)$, not directly to raw pixels. This stage searches for a candidate effective mesoscale law describing the evolution of $c$. The candidate term library is
\begin{equation}
\begin{aligned}
\Theta(c)=\{&
1,c,c^2,\partial_x c,\partial_y c,\nabla^2 c,
|\nabla c|^2,\sigma,h_{\mathrm{env}}, \\
&c\,\sigma,c\,h_{\mathrm{env}},
\nabla\cdot(D(c)\nabla c),\ldots
\}
\end{aligned}
\label{eq:thetalib}
\end{equation}
In Eq.~\eqref{eq:thetalib}, $\sigma$ is a measured or derived stress-, preload-, or contact-related descriptor, $h_{\mathrm{env}}$ denotes a measured exposure/environment descriptor, and $D(c)$ is a possible state-dependent diffusivity. The library contains scalar candidate terms for the scalar evolution equation of $c$, including transport, nonlinear severity terms, and measured coupling variables. It excludes unconstrained latent terms in the first discovery mode.

Two discovery modes are distinguished. The physics-readable mode uses only $c$, its spatial derivatives, time, measured exposure variables, and measured preload/contact descriptors:
\begin{equation}
\frac{\partial c}{\partial t}=\mathcal{N}_{r}\left(c,\nabla c,\nabla^2c,\sigma,h_{\mathrm{env}}\right)
\label{eq:pde_readable}
\end{equation}
The latent-predictive mode permits selected latent descriptors when they improve prediction, but such terms are penalized more strongly and interpreted only as predictive terms:
\begin{equation}
\frac{\partial c}{\partial t}=\mathcal{N}_{\ell}\left(c,\nabla c,\nabla^2c,\sigma,h_{\mathrm{env}},\vect{z}(\vect{s},t)\right)
\label{eq:pde_latent}
\end{equation}
Equations~\eqref{eq:pde_readable} and \eqref{eq:pde_latent} make the interpretability boundary explicit. A compact equation containing $\vect{z}$ may be useful for forecasting, but not interpreted as a mechanistic electrochemical law.

A genetic algorithm encodes candidate equations as chromosomes whose genes select terms from $\Theta(c)$. Selection, crossover, and mutation are used to evolve the population. The fitness of a candidate model $\mathcal{M}$ is defined as
\begin{equation}
J(\mathcal{M})=E_{\mathrm{weak}}(\mathcal{M})+\lambda_s|\mathcal{M}|+\lambda_{\mathrm{phys}}P(\mathcal{M})+\lambda_zZ(\mathcal{M})
\label{eq:gafitness}
\end{equation}
where $E_{\mathrm{weak}}$ is the weak-form PDE residual, $|\mathcal{M}|$ is the number or symbolic complexity of retained terms, $P(\mathcal{M})$ penalizes physical violations such as negative severity or dimensionally implausible terms, and $Z(\mathcal{M})$ penalizes latent-variable dependence. The weights $\lambda_s$, $\lambda_{\mathrm{phys}}$, and $\lambda_z$ control parsimony, physical plausibility, and latent-term use. Genetic and symbolic PDE-discovery methods are appropriate when the candidate equation is not guaranteed to lie in a fixed polynomial library \cite{xu2020dlga,chen2022sga,cohen2024pigp}.

For noisy reconstructed fields, the weak-form residual is preferred:
\begin{equation}
E_{\mathrm{weak}}(\mathcal{M})=\sum_{\ell}\left|\int_{0}^{T}\int_{\surf}\left(\partial_t c-\mathcal{N}_{\mathcal{M}}(c)\right)\varphi_{\ell}(\vect{s},t)\,\diff\vect{s}\,\diff t\right|^2
\label{eq:weakresidual}
\end{equation}
where $\varphi_{\ell}$ are test functions and $\mathcal{N}_{\mathcal{M}}$ is the differential operator represented by candidate model $\mathcal{M}$. Weak-form PDE identification reduces the effect of pointwise differentiation noise \cite{messenger2021wsindy,stephany2024weakpdelearn}.


\subsection{Physics-Informed Prediction Model}
The discovered candidate law is then used as a constraint inside the prediction model. The prediction layer separates corrosion evolution from functional behavior. The corrosion branch estimates the evolution of $c$, $p$, and $h_s$ over the surface. The functional branch maps descriptors extracted from these fields to selected functional indicators. This separation is necessary because a surface corrosion PDE does not automatically produce torque-angle or force-displacement curves.

The corrosion branch is represented by a neural field,
\begin{equation}
\left[\hat{c},\hat{p},\hat{h}_s\right]=u_{\theta}(\vect{s},t,\vect{q},\vect{e})
\label{eq:neuralfield}
\end{equation}
where $u_{\theta}$ is the physics-informed neural field, $\hat{c}$ is predicted corrosion severity, $\hat{p}$ is predicted pit probability or density, and $\hat{h}_s$ is predicted surface-height degradation. The PDE residual for the corrosion field is
\begin{equation}
\mathcal{R}_{\theta}=\partial_t\hat{c}-\mathcal{N}_{\mathcal{M}}\left(\hat{c},\nabla\hat{c},\nabla^2\hat{c},\sigma,h_{\mathrm{env}}\right)
\label{eq:pinnresidual}
\end{equation}
In Eq.~\eqref{eq:pinnresidual}, $\mathcal{R}_{\theta}$ measures violation of the selected candidate PDE $\mathcal{M}$. The covariates $\sigma$ and $h_{\mathrm{env}}$ are measured directly when available or derived from $\vect{q}$ and $\vect{e}$. PINNs are used here to combine field observations with the PDE residual \cite{raissi2019pinn}.

\begin{figure*}[t]
    \centering
    \includegraphics[width=0.8\textwidth, trim=0cm 0cm 0cm 0cm, clip, keepaspectratio]{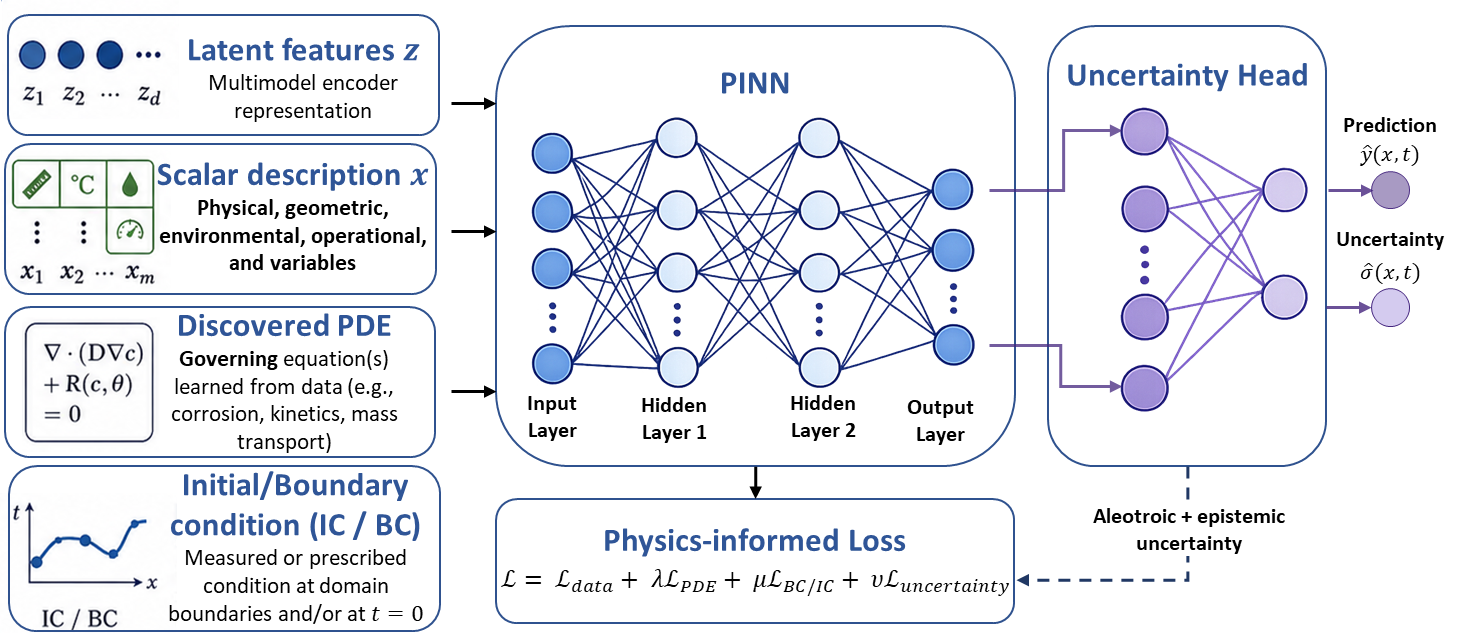}
    \caption{Coupled physics-informed prediction model. The corrosion branch is constrained by the candidate PDE, while the functional branch maps corrosion descriptors to selected indicators such as friction coefficient and, when measured, preload loss.}
    \label{fig:pinn}
\end{figure*}

The first functional target is the friction coefficient,
\begin{equation}
\begin{aligned}
\hat{\mu}_{i,k}=g_{\phi}\big(&
\bar{c}_{i,k}, A_{r,i,k}, \rho_{p,i,k},
\bar{h}_{s,i,k}, R_{a,i,k}, R_{z,i,k}, \\
&F_{p,i,0}, \vect{q}_{i}, \vect{e}_{i,k}
\big)
\end{aligned}
\label{eq:functionalmap}
\end{equation}
Here $g_{\phi}$ is the functional-behavior model, $\hat{\mu}_{i,k}$ is the predicted friction coefficient, $\bar{c}_{i,k}$ is the average or region-specific corrosion severity, $A_{r,i,k}$ is rust-area fraction, $\rho_{p,i,k}$ is pit density, $\bar{h}_{s,i,k}$ is mean surface-height degradation, $R_a$ and $R_z$ are roughness descriptors, $F_{p,i,0}$ is the initial preload or clamp-force condition for specimen $i$, and $\vect{q}_{i}$ and $\vect{e}_{i,k}$ are joint and exposure descriptors. Preload loss $\Delta F_p$ can be modeled by the same structure with clamp-force data.
The total learning objective is
\begin{equation}
\loss=\loss_{c}+\alpha\|\mathcal{R}_{\theta}\|^2+\beta\loss_{\mathrm{BC/IC}}+\gamma\loss_{\mathrm{func}}+\delta\loss_{\mathrm{unc}}
\label{eq:totalloss}
\end{equation}
where $\loss_c$ is the calibration loss from Eq.~\eqref{eq:calibrationloss}, $\|\mathcal{R}_{\theta}\|^2$ is the PDE residual loss from Eq.~\eqref{eq:pinnresidual}, $\loss_{\mathrm{BC/IC}}$ enforces boundary and initial conditions, $\loss_{\mathrm{func}}$ fits the selected measured functional indicators, and $\loss_{\mathrm{unc}}$ calibrates uncertainty. The weights $\alpha$, $\beta$, $\gamma$, and $\delta$ balance physics, boundary information, functional prediction, and uncertainty. Loss imbalance and boundary-condition errors are known training risks in PINNs \cite{wang2022pinnfail,wang2024expert}.
Gradient-enhanced residuals and exact boundary imposition are relevant when derivative fidelity or domain boundaries are critical \cite{yu2022gpinn,sukumar2022bc}.
Uncertainty can be represented through ensembles or Bayesian PINNs when noisy measurements are expected \cite{yang2021bpinn}.
Figure~\ref{fig:pinn} shows the coupled prediction architecture.

\subsection{Predicted Outputs and Engineering Decision Support}
The final step translates model predictions into quantities that can support inspection decisions. The model outputs are interpreted at two levels. The first level consists of predicted degradation fields and scalar condition indicators, including corrosion severity, rust coverage, pit density, roughness-related descriptors, and uncertainty. The second level consists of selected functional indicators, primarily friction coefficient and secondarily preload loss when clamp-force data are available. Full torque-angle or force-displacement curve prediction is treated as a later extension because it requires additional mechanical coupling assumptions.

For circularity-oriented inspection, the useful output is not only the predicted value but also its confidence and traceability. A predicted friction change with high uncertainty may indicate the need for further inspection rather than immediate reuse or refurbishment. A stable low-severity field with acceptable predicted friction may support reuse. A moderate condition with recoverable function may support refurbishment. Severe corrosion, high uncertainty, or predicted functional degradation may support recycling or disassembly. Thus, the framework provides a structured path from observable degradation evidence to decision-relevant functional indicators.

\section{VALIDATION STRATEGY}
Validation is defined as a set of linked checks rather than a single
end-to-end score. Each check tests one interface in the proposed workflow:
multimodal field construction, uncertainty propagation, PDE discovery, and
functional prediction. To avoid leakage, validation should use specimen-level
splits; patches, repeated views, and repeated measurements from the same
specimen should remain in the same split. Temporal rollout should be evaluated
on exposure intervals not used for fitting or model selection.
First, the multimodal field construction must be verified. Cross-modal
registration is evaluated on held-out specimens using target registration
error, inlier ratio, and deformation smoothness. Registration error must be
smaller than the spatial feature scale used for gradients, such as the relevant
pit diameter or roughness length scale. The reconstructed fields are then
compared with independent measurements: rust masks with Dice or intersection
over union (IoU), pit maps
with pit-count or pit-density error, and surface-height fields with
profilometry error. The calibrated corrosion-severity field
$c(\vect{s},t)$ must also remain consistent with independent degradation
evidence, such as roughness, pit-density, or material-loss measurements.

Second, error propagation is tested before interpreting any discovered PDE.
Registration transforms, segmentation masks, roughness values, exposure times,
and calibration inputs are perturbed within plausible uncertainty ranges. Term
stability is measured as
\begin{equation}
    S_{\nu}=\frac{1}{B}\sum_{b=1}^{B}
    \mathbb{I}\left(T_{\nu}\in\mathcal{M}^{(b)}\right)
    \label{eq:termstability}
\end{equation}
where $S_{\nu}$ is the stability score of candidate term $T_{\nu}$, $B$ is the
number of perturbation or bootstrap runs, $\mathcal{M}^{(b)}$ is the selected
equation in run $b$, and $\mathbb{I}(\cdot)$ is the indicator function. Terms
with low stability are treated as artifacts of reconstruction or noise rather
than as components of a candidate effective law.

Third, PDE discovery is tested on synthetic corrosion-inspired fields with
known governing structure before real-data interpretation. This step reports
term precision, term recall, coefficient error, and rollout stability.
PDE-FIND, weak-form Sparse Identification of Nonlinear Dynamics (SINDy),
manually specified candidate laws, and genetic
symbolic discovery are compared under the same noise conditions
\cite{rudy2017pdefind,messenger2021wsindy,xu2020dlga}. After this benchmark,
candidate laws discovered from experimental fields are interpreted as
candidate effective laws only if they remain stable under perturbation and
improve held-out temporal rollout, rather than only reducing training
residuals.

Fourth, specimen-level functional prediction is evaluated. The primary test is
whether image-derived corrosion fields improve prediction of $\mu$ beyond
roughness descriptors, exposure time, and joint configuration alone. When
clamp-force data are available, the same validation logic is applied to preload
loss $\Delta F_p$. Baselines include image-only regression, scalar-only
regression, a no-PDE neural predictor, a manually specified PDE constraint,
PDE-FIND, and weak-form SINDy. The framework is falsified if
$c(\vect{s},t)$ does not correlate with independent degradation measurements,
if discovered PDE terms are unstable under perturbation, if the PDE improves
training loss but not held-out rollout, or if the image-derived fields do not
improve specimen-level prediction of the selected functional indicators.

\section{DISCUSSION}
The discussion focuses on what the framework
proposes conceptually, how it can support inspection decisions, and which conditions must be satisfied before empirical claims about corrosion evolution or bolted-joint function can be made.

\subsection{Implications}
The framework changes the role of images in corrosion assessment. Images are not treated as final labels or as direct measurements of a governing law. They are treated as evidence for calibrated degradation fields that can be tested
against independent measurements and used in physics-informed prediction. This
distinction is important because it prevents visual texture from being
interpreted as physics without calibration.

The framework also clarifies the meaning of PDE discovery in
inspection-oriented corrosion analysis. A discovered equation over
$c(\vect{s},t)$ is a candidate effective mesoscale law. It can support
prediction if it is stable under perturbation and improves held-out rollout,
but it is not a complete electrochemical mechanism unless the required
electrochemical variables and boundary conditions are measured. This
interpretation keeps the method scientifically bounded while preserving the
value of symbolic structure.

\subsection{Advantages for Inspection and Circularity}
The main practical advantage is that the framework begins with feasible
inspection data. Camera images, microscopy, roughness measurements, and
selected torque/clamp-force tests are more realistic for returned products than
full electrochemical characterization. A second advantage is traceability:
each functional prediction is linked to surface observations, reconstructed
fields, and model assumptions. A third advantage is interpretability:
calibrated field definitions and symbolic PDE discovery provide more insight
than black-box image regression alone. A fourth advantage is uncertainty
awareness, which is essential when degradation histories are unknown.

For circularity-oriented engineering, the framework supports a decision chain
from observed condition to functional indicator to action. Reuse can be
supported when degradation and predicted functional change are low and
uncertainty is acceptable. Refurbishment can be supported when degradation is
moderate but the functional indicator can be restored. Recycling or
disassembly can be supported when degradation is severe, uncertainty is high,
or predicted function is unacceptable. The framework therefore provides a
methodological basis for using inspection evidence in R-strategy decisions.

\subsection{Limitations and Failure Conditions}
The framework is data-intensive. Time-dependent PDE discovery requires
repeated or re-localizable observations of the same surface regions.
Cross-modal registration errors can introduce artificial gradients. The
surrogate field $c$ may be predictive without being physically identifiable if
it does not correlate with independent roughness, pit, or material-loss
measurements. PINN training may become unstable if loss terms are poorly
balanced or boundary conditions are not defined. Genetic symbolic discovery can overfit noisy fields without parsimony penalties and perturbation-stability checks. These limitations define the conditions under which the proposed framework is restricted. In particular, it is a structured methodology ready for validation, but it must be tested on specimen-level splits with repeated exposure intervals and independent functional measurements.

\section{CONCLUSIONS AND FUTURE WORK}
This paper formulated a physics-informed learning framework for linking
microscopic surface degradation to bolted-joint functional behavior. The
framework follows a reproducible sequence: define the corrosion failure mode,
generate multimodal inspection and functional-response data, structure the data
into descriptors, learn cross-scale similarity between microscopic and camera
observations, construct a calibrated corrosion-severity field, discover
candidate effective PDEs, and couple corrosion evolution to selected
functional indicators. The first functional target is friction coefficient,
while preload loss is introduced when clamp-force data are available.

The contribution is the methodological architecture connecting image-based
degradation information to physically constrained prediction for degraded
bolted joints. The framework is intentionally bounded: discovered PDEs are
interpreted as candidate effective laws over calibrated degradation fields, not
as complete electrochemical mechanisms. Future work will implement the
framework on a dedicated experimental dataset with specimen-level splits,
repeated exposure intervals, paired microscopy and camera observations,
roughness measurements, and torque/clamp-force tests. Synthetic corrosion
benchmarks will be used before interpreting real-data PDEs. The central
empirical question is whether calibrated image-derived corrosion fields
improve held-out prediction of selected functional indicators while producing
stable candidate effective laws.
\section*{Acknowledgements}
This work was funded by the Deutsche Forschungsgemeinschaft (DFG, German Research Foundation) – "Functional modeling of bolted joints under uncertain product conditions for remanufacturing" - with the project number 525034540. During the preparation of this work, the authors used Grammerly and DeepL in order to improve the structure, readability, and phrasing of the manuscript. After using these tools, the authors reviewed and edited the content as needed and take full responsibility for the content of the publication.
\bibliographystyle{asmeconf}
\bibliography{references_revised_v8}

\end{document}